# Fusion of Multispectral Data Through Illumination-aware Deep Neural Networks for Pedestrian Detection

Dayan Guan[b], Yanpeng Cao[a,b,∗], Jiangxin Yang[a,b], Yanlong Cao[a,b], Michael Ying Yang[c]

[a]*State Key Laboratory of Fluid Power and Mechatronic Systems, School of Mechanical Engineering, Zhejiang University, Hangzhou, China*
[b]*Key Laboratory of Advanced Manufacturing Technology of Zhejiang Province, School of Mechanical Engineering, Zhejiang University, Hangzhou, China*
[c]*Scene Understanding Group, University of Twente, Hengelosestraat 99, 7514 AE Enschede, The Netherlands*

## Abstract

Multispectral pedestrian detection has received extensive attention in recent years as a promising solution to facilitate robust human target detection for around-the-clock applications (*e.g.* security surveillance and autonomous driving). In this paper, we demonstrate illumination information encoded in multispectral images can be utilized to significantly boost performance of pedestrian detection. A novel illumination-aware weighting mechanism is present to accurately depict illumination condition of a scene. Such illumination information is incorporated into two-stream deep convolutional neural networks to learn multispectral human-related features under different illumination conditions (daytime and nighttime). Moreover, we utilized illumination information together with multispectral data to generate more accurate semantic segmentation which are used to boost pedestrian detection accuracy. Putting all of the pieces together, we present a powerful framework for multispectral pedestrian detection based on multi-task learning of illumination-aware pedestrian detection and semantic segmentation. Our proposed method is trained end-to-end using a well-designed multi-task loss function and outperforms state-of-the-art approaches on KAIST multispectral pedestrian dataset.

## 1. INTRODUCTION

Pedestrian detection is a popular research topic within the field of computer vision in the past decades [29, 5, 8, 11, 10, 4, 41]. Given images captured in various realworld surveillance situations, pedestrian detection solution is required to generate bounding boxes to accurately locate individual pedestrian instances. It provide an important functionality to facilitate a board range of human-centric applications, such as video surveillance [36, 1, 25] and autonomous driving [37, 24, 39].

Although significant improvements have been accomplished in recent years, developing a robust pedestrian detection solution which is ready for practical applications still remains a challenging task. It is noticed that most existing pedestrian detectors are trained using visible information alone thus their performances are sensitive to changes of illumination, weather and occlusions [18]. To overcome the aforementioned limitations, many research works have been focused on the development of multispectral pedestrian detection solutions to facilitate robust human target detection for around-the-clock application [22, 21, 34, 28, 16, 13]. The underlying intuition is that multispectral images (*e.g.* visible and thermal) provide complementary information about objects of interest and effective fusion of such data can lead to more robust and accurate detection results.

In this work, we present a framework for learning multispectral human-related characteristics under various illumination conditions (daytime and nighttime) through the proposed illumination-aware deep neural networks. We observed that multispectral pedestrian instances exhibit significantly different human-related characteristics under day and night illumination conditions as illustrated in Figure 1, thus using multiple built-in sub-networks, each of which specializes in capturing illumination-specific visual patterns, provides an effective solution to handle substantial intra-class variance cased by various illumination conditions for more robust target detection. Illumination information can be robustly estimated based on multispectral data and is further infused into multiple illumination-aware sub-networks to learn multispectral semantic feature maps for robust pedestrian detection and semantic segmentation under different illumination conditions. Given a pair of multispectral images captured during daytime, our proposed illumination-aware weighting mechanism adaptively assigns a high weight for day-illumination sub-networks (pedestrian detection and semantic segmentation) to learn human-related characteristics in daytime. In comparison, multispectral images of a nighttime scene are utilized to generate night-illumination features. We provide an illustration of how this illumination-aware weighting mechanism works in Figure 2. The final detection results are generated by fusion the outputs of multiple illumination-aware sub-networks and remain robust to large variance in scene illumination changes. The contributions of this work are as follows.

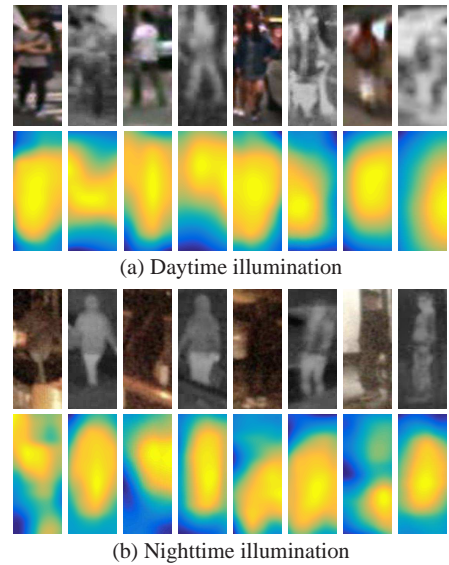

(a) Daytime illumination

(b) Nighttime illumination

Figure 1: Characteristics of multispectral pedestrian instances captured in (a) daytime and (b) nighttime scenes. The first rows in (a) and (b) show the multispectral picture of pedestrian instances. The second rows in (a) and (b) show the feature map visualizations of the corresponding pedestrian instances. The feature maps of visible and thermal images are generated using the deep neural region proposal networks [38] well-trained in their correspondidng channels. Notice that multispectral pedestrian instances exhibit significantly different human-related characteristics under daytime and nighttime illumination conditions.

∗Corresponding author

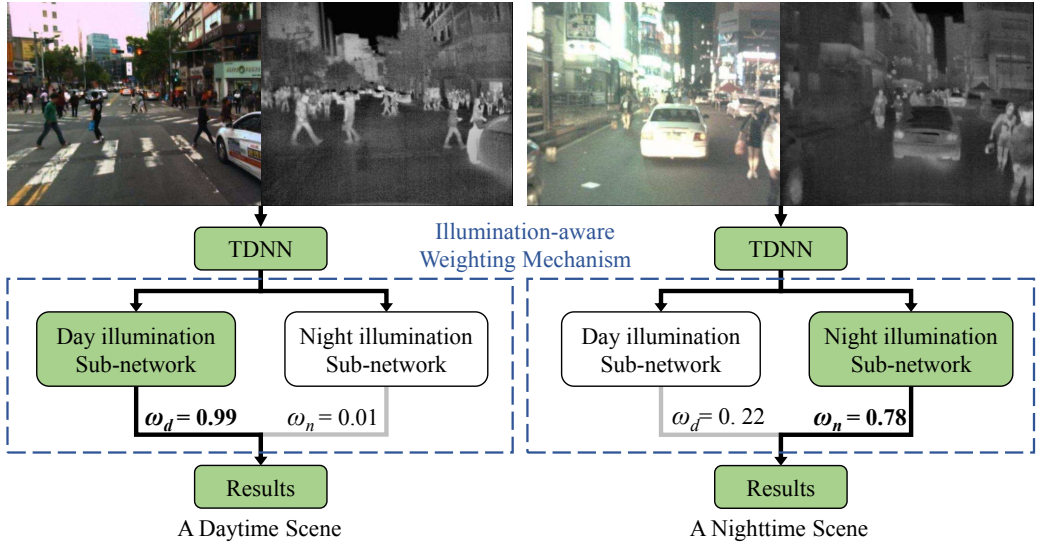

Figure 2: Illustration of the illumination-aware weighting mechanism. Given a pair of aligned visible and thermal images, two-stream deep neural networks (TDNN) generate multispectral semantic feature maps. Day-illumination sub-networks and night-illumination ones utilize the multispectral semantic feature maps for pedestrian detection and semantic segmentation under different illumination conditions. The final detection results are generated by fusing the outputs of multiple illumination-aware sub-networks.

Firstly, we demonstrate that illumination condition of a scene can be robustly determined through an architecture of fully connected neural networks by considering multispectral semantic features and the estimated illumination information provides useful information to boost performance of pedestrian detection.

Secondly, we incorporate an illumination-aware mechanism into two-stream deep convolutional neural networks to learn multispectral human-related features under different illumination conditions (daytime and nighttime). To the best of our knowledge, this is the first attempt to explore illumination information for training multispectral pedestrian detector.

Thirdly, we present a complete framework for multispectral pedestrian detection based on multi-task learning of illumination-aware pedestrian detection and semantic segmentation which is trained end-to-end using a well-designed multi-task loss. Our method achieves lower miss rate and faster runtime compared with the state-of-the-art multispectral pedestrian detectors [16, 18, 19].

The remainder of the paper is organized as follows. We review some existing solutions for multispectral pedestrian detection in Section 2. The details of our proposed illumination-aware deep neural networks are presented in Section 3. An extensive experimental comparison of methods for multispectral pedestrian detection is provided in Section 4, and Section 5 concludes this paper.

## 2. Related Work

Pedestrian detection approaches using visible and multispectral images are closely related to our work. We present a review of the latest researches on these topics below.

**Visible Pedestrian Detection:** A large variety of methods have been presented to perform pedestrian detection using visible information. Integrate Channel Features (ICF) pedestrian detector presented by Piotr *et al.* is based on feature pyramids and boosted classifiers [6]. Its performance has been further improved through multiple techniques including ACF[7], LDCF[27], and Checkerboards[40] etc. Recently, DNNs-based approaches for object detection [12, 31, 15] have been adopted to improve the performance of pedestrian detection. Li *et al.* [23] presented a Scale-aware deep network framework in which a large-size sub-network and a small-size one are combined into a unified architecture to depict unique pedestrian features at different scales. A unified architecture of multi-scale deep neural networks is presented by Cai *et al.* [3] to combine complementary scale-specific detectors together, thus it provides a number of receptive fields to match objects of different scales. Zhang *et al.* [38] made use of high-resolution convolutional feature maps for classification and presented an effective pipeline for pedestrian detection using region proposal networks(RPN) followed by boosted forests. Mao *et al.* [26] proposed a novel network architecture to jointly learn pedestrian detection as well as the given extra feature. This multi-task training scheme is able to utilize the information of given features and improve detection performance without extra inputs in inference. Brazil*et al.* [2] developed a segmentation infusion network to boost pedestrian detection accuracy with the joint supervision on semantic segmentation and pedestrian detection. It is proved that weakly annotated boxes provide sufficient information to achieve considerable performance gains.

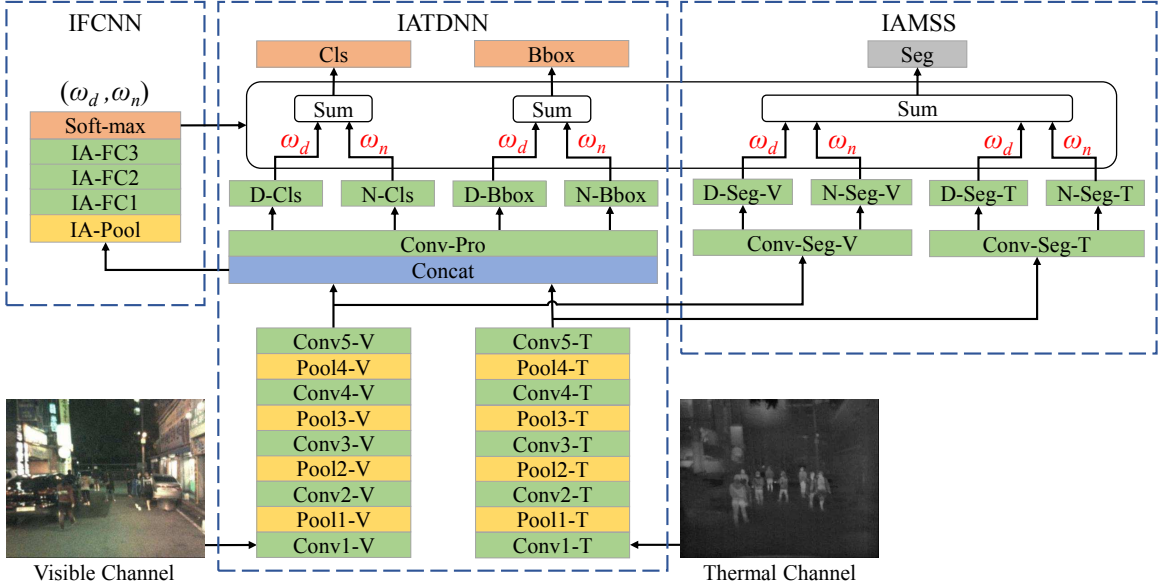

Figure 3: The architecture of our proposed illumination-aware multispectral deep neural networks (IATDNN+IASS). Note that green boxes represent convolutional and fully-connected layers, yellow boxes represent pooling layers, blue boxes represent fusion layers, gray boxes represent segmentation layers, and orange boxes represent output layers. Best viewed in color.

**Multispectral Pedestrian Detection:** Multispectral images provide complementary information about objects of interest, thus pedestrian detectors trained using multi-modal data sources produce robust detection results. A large-size multispectral pedestrian dataset (KAIST) is presented by Hwang *et al.* [16]. With well-aligned visible and thermal image pairs with dense pedestrian annotations, the author proposed a new multispectral aggregated features (ACF+T+THOG) to process color-thermal image pairs and applied a boosted decision trees (BDT) for target classification. Wagner *et al.* [35] presented the first application of DNNs for multispectral pedestrian detection and evaluated the performance of two decision networks (early-fusion and late-fusion). These decision networks verify pedestrian candidates generated by ACF+T+THOG [16] to achieve more accurate detection results. Liu *et al.* [18] investigate how to utilize Faster R-CNN [31] for multispectral pedestrian detection task and designed four ConvNet fusion architectures in which two-branch ConvNets are integrated at different DNNs stages. The optimal architecture is the Halfway Fusion model that merge two-branch ConvNets using the middle-level convolutional features. König *et al.* [19] modified the architecture of RPN + BDT [38] to build Fusion RPN + BDT for multispectral pedestrian detection. The Fusion RPN merges the two-branch RPN on the middle-level convolutional features and achieves the state-of-the-art performance on KAIST multispectral dataset. Our approach differs from the above methods distinctly by developing a framework to learn multispectral human-related features under different illumination conditions (daytime and nighttime) through the proposed illumination-aware multispectral deep neural networks. To the best of our knowledge, this is the first attempt to explore illumination information to boost multispectral pedestrian detection performances.

## 3. Our Approach

### 3.1. Overview of Proposed Model

The architecture of illumination-aware multispectral deep neural networks is illustrated in Figure 3. It consists of three integrated processing modules including illumination fully connected neural networks (IFCNN), illumination-aware two-stream deep convolutional neural networks (IATDNN), and illumination-aware multispectral semantic segmentation (IAMSS). Given aligned visible and thermal images, IFCNN computes the illumination-aware weights to determine whether it is is daytime scene or night one. Through the proposed illumination-aware mechanism, IATDNN and IASS make use of multi sub-networks to generate detection results (classification scores - Cls and bounding boxes - Bbox) and segmentation masks (Seg). For instance, IATDNN employ two individual classification sub-networks (D-Cls and N-Cls) for human classification under day and night illuminations. Cls, Bbox and Seg results of each sub-networks are combined to generate the final output through a gate function which is defined over the illumination condition of the scene. Our proposed method is trained end-to-end based on multi-task learning of illumination-aware pedestrian detection and semantic segmentation.

### 3.2. Illumination Fully Connected Neutral Networks (IFCNN)

As shown in Figure 3, a pair of visible and thermal images are passed into the first five convolutional layers and pooling ones of two-stream deep convolutional neural networks (TDNN) [19] to extract semantic feature maps in individual channels. Note that each stream of feature extraction layers in TDNN (Conv1-V to Conv5-V in the visible stream and Conv1-T to Conv5-T in the thermal stream) uses Conv1-5 from VGG-16 [33] as the backbone. Then feature maps from two channels are fused to generate the two-stream feature maps (TSFM) through a concatenate layer (Concat). TSFM is utilized as the input of IFCNN to compute illumination-aware weights $\omega_d$ and $\omega_n = (1 - \omega_d)$ which determine the illumination condition of a scene.

The IFCNN consist of a pooling layer (IA-Pool), three fully connected layers (IA-FC1, IA-FC2, IA-FC3), and the soft-max layer (Soft-max). Similar to the spatial pyramid pooling (SPP) layer which removes the fixed-size constraint of the network [14], IA-Pool resizes the features of TSFM to a fixed-length figure maps (7×7) using bilinear interpolation and generates fixed-size outputs for the fully connected layers. The number of channels in IA-FC1, IA-FC2, IA-FC3 are empirically set to 512, 64, 2 respectively. Soft-max is the final layer of IFCNN. The outputs of Soft-max are $\omega_d$ and $\omega_n$. We define the illumination error term $L_I$ as

$$L_I = -\hat{\omega}_d \cdot \log(\omega_d) - \hat{\omega}_n \cdot \log(\omega_n), \qquad (1)$$

where $\omega_d$ and $\omega_n = (1 - \omega_d)$ are the estimated illumination weights for day and night scenes, $\hat{\omega}_d$ and $\hat{\omega}_n = (1 - \hat{\omega}_d)$ are the illumination labels. If the training images are captured under daytime illumination conditions, we set $\hat{\omega}_d = 1$, otherwise $\hat{\omega}_d = 0$.

### 3.3. Illumination-aware Two-Stream Deep Convolutional Neutral Networks (IATDNN)

The architecture of IATDNN is designed based on the two-stream deep convolutional neural networks (TDNN) [19]. Region proposal networks (RPN) model [38] is adopted in IATDNN due to its superior performance for pedestrian detection. Given a single input image, RPN outputs a number of bounding boxes associated with confident scores to generate pedestrian proposals through classification and bounding box regression. As shown in Figure 4(a), a 3×3 convolutional layer (Conv-Pro) is attached after Concat layer with two sibling 1×1 convolutional layers (Cls and Bbox) for classification and bounding box regression respectively. TDNN model provides an effective framework to utilize two-stream feature maps (TSFM) for robust pedestrian detection.

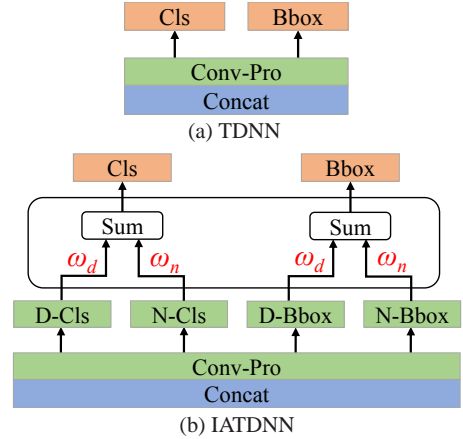

Figure 4: The comparison of TDNN and IATDNN architectures. Note that $\omega_d$ and $\omega_n$ is , green boxes represent convolutional and fully-connected layers, yellow boxes represent pooling layers, blue boxes represent fusion layers, and orange boxes represent output layers. Best viewed in color.

We further incorporate illumination information into TDNN to generate classification and regression results for various illumination conditions. Specifically, IATDNN contains four sub-networks (D-Cls, N-Cls, D-Bbox, and N-Bbox) to produce illumination-aware detection results as shown in Figure 4(b). D-Cls and N-Cls calculate classification scores under day and night illumination conditions while D-Bbox and N-Bbox generate bounding boxes for daytime and nighttime scenes respectively. The outputs of these sub-networks are combined using the illuminating weights calculated in IFCNN to produce final detection results. The detection loss term $L_{DE}$ is defined as

$$L_D = \sum_{i \in S} L_{cls}(c_i^f, \hat{c}_i) + \lambda_{bb} \cdot \hat{c}_i \cdot \sum_{i \in S} L_{bbox}(b_i^f, \hat{b}_i), \qquad (2)$$

where $L_{DE}$ is the combination of classification loss $L_{cls}$ and regression loss $L_{bbox}$, $\lambda_{bb}$ defines the regularization parameter between them (we set $\lambda_{bb} = 5$ according to the method presented by Zhang *et al.* [38]), $S$ defines the set of training samples in one mini-batch. A training sample is considered as a positive if its Intersection-over-Union (IoU) ratio with one ground truth bounding box is greater than 0.5, and otherwise negative. We set training label $\hat{c}_i = 1$ for positive samples and $\hat{c}_i = 0$ for negative ones. For each positive sample, its bounding box is set to $\hat{b}_i$ for computing the bounding box regression loss. In Eq. 2, the classification loss term $L_{cls}$ is defined as

$$L_{cls}(c_i^f, \hat{c}_i) = -\hat{c}_i \cdot \log(c_i^f) - (1 - \hat{c}_i) \cdot \log(1 - c_i^f), \qquad (3)$$

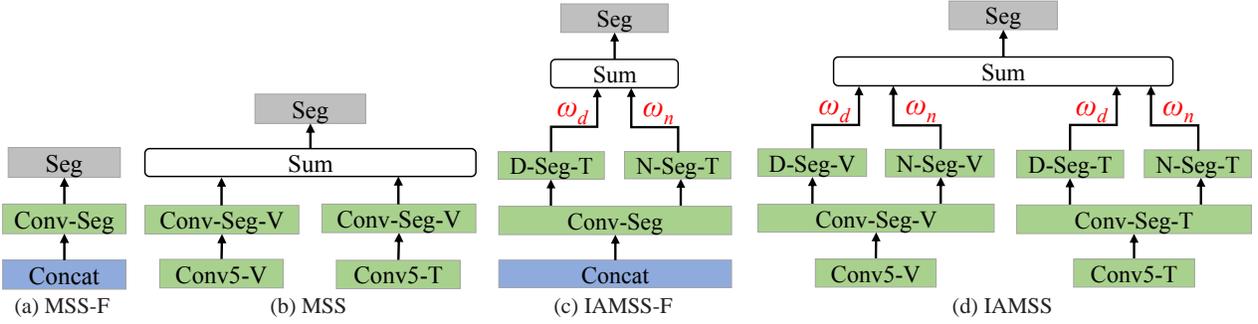

Figure 5: The comparison of MSS-F, MSS, IAMSS-F and IAMSS architectures. Note that green boxes represent convolutional layers, blue boxes represent fusion layers, and gray boxes represent segmentation layers. Best viewed in color.

and the regression loss term $L_{bbox}$ is defined as

$$L_{bbox}(b_i^f, \hat{b}_i) = \sum smooth_{L_1}(b_{ij}^f, \hat{b}_{ij}) \quad (4)$$

where $c_i^f$ and $b_i^f$ are the predicted classification score and bounding box respectively, and the $L1$ loss function $smooth_{L_1}$ is defined in [12] to learn the transformation mapping between $b_i^f$ and $\hat{b}_i^f$. In IATDNN, $c_i^f$ is calculated as the weighted sum of day-illumination classification score $c_i^d$ and night-illumination classification score $c_i^n$ as

$$c_i^f = \omega_d \cdot c_i^d + \omega_n \cdot c_i^n, \quad (5)$$

and $b_i^f$ is the illumination weighted combination of two bounding boxes $b_i^d$ and $b_i^n$ predicted by D-Bbox and N-Bbox sub-networks respectively as

$$b_i^f = \omega_d \cdot b_i^d + \omega_n \cdot b_i^n. \quad (6)$$

Through the above illumination weighting mechanism, the day-illumination sub-networks (classification and regression) will be given a high priority to learn human-related characteristics in daytime scene. On the other hand, multispectral feature maps of a nighttime scene are utilized to generate reliable detection results under night-illumination conditions.

### 3.4. Illumination-aware Semantic Segmentation (IASS)

Recently, semantic segmentation masks have been successfully used as strong cues to improve performance of single channel based object detection [15, 2]. The simple box-based segmentation masks provide additional supervision to guide features in shared layers become more distinctive for the downstream pedestrian detector. In this paper we incorporate the semantic segmentation scheme with two-stream deep convolutional neutral networks to enable simultaneous pedestrian detection and segmentation on multispectral images.

Given information from two multispectral channels (visible and thermal), fusion at different stages (feature-stage and decision-stage) would lead to different segmentation results. Therefore, we hope to investigate what is the best fusion architecture for multispectral segmentation task. To this end, we design two multispectral semantic segmentation architectures that perform fusions at different stages, denoted as feature-stage multispectral semantic segmentation (MSS-F) and decision-stage multispectral semantic segmentation (MSS). As shown in Figure 5(a)-(b), MSS-F firstly concatenates the feature maps from Conv5-V and Conv5-T and then applies a common Conv-Seg layer to produce segmentation masks. In comparison, MSS applies two convolutional layer (Conv-Seg-V and Conv-Seg-T) to produce different segmentation maps for individual channels and then combine two-stream outputs to generate the final segmentation masks.

Moreover, we hope to investigate whether the performance of semantic segmentation can be boosted by considering illumination condition of the scene. Based on MSS-F and MSS architectures, we design two more illumination-aware multispectral semantic segmentation architectures (IAMSS-F and IAMSS). As shown in Figure 5(c)-(d), two segmentation sub-networks (D-Seg and N-seg) are employed to generate illumination-aware semantic segmentation results. Note that IAMSS-F contains two sub-networks and IAMSS contains four sub-networks. The outputs of these sub-networks are fused through the illumination weighting mechanism to generate the multispectral semantic segmentation using the illuminating weights predicted by IFCNN. In Section 4, we provide evaluation results of these four differnt multispectral segmentation architectures.

Here we define the segmentation loss term as

$$L_S = \sum_{i \in C} \sum_{j \in B} [-\hat{s}_j \cdot \log(s_{ij}^f) - (1 - \hat{s}_j) \cdot \log(1 - s_{ij}^f)], \quad (7)$$

where $s_{ij}^f$ is the predicted segmentation mask, $C$ are segmentation streams ( MSS-F and IAMSS-F contain only one segmentation stream while MSS and IAMSS contain two streams), $B$ are box-based segmentation training samples in one mini-batch. If the sample is within a ground truth bounding box, we set $\hat{s}_j = 1$, otherwise $\hat{s}_j = 0$. In illumination-aware multispectral semantic segmentation architectures IAMSS-F and IAMSS, $s_{ij}^f$ is the illumination weighted combination of two segmentation masks $s_{ij}^d$ and $s_{ij}^n$ predicted by D-Seg and N-Seg sub-networks respectively as

$$s_{ij}^f = \omega_d \cdot s_{ij}^d + \omega_n \cdot s_{ij}^n. \quad (8)$$

To perform multi-task learning of illumination-aware pedestrian detection and semantic segmentation, we combine the loss terms defined in Eq. 1, 2, 7 and our final multi-task loss function becomes

$$L_{I+D+S} = L_D + \lambda_{ia} \cdot L_I + \lambda_{sm} \cdot L_S \quad (9)$$

where $\lambda_{ia}$ and $\lambda_{sm}$ are the trade-off coefficient of loss term $L_I$ and $L_S$ respectively. We set $\lambda_{ia} = 1$ and $\lambda_{sm} = 1$ according to the method presented by Brazil *et al.* [2]. We make use of this loss function to jointly train illumination-aware multispectral deep neural networks.

## 4. Experiments

### 4.1. Experimental Setup

**Datasets:** Our experiments are conducted using the public KAIST multispectral pedestrian benchmark [16]. In total, KAIST training dataset contains 50,172 aligned color-infrared image pairs captured at various urban locations and under different lighting conditions with dense annotations. We sample images every 2 frames and obtain 25,086 training images following the method presented by König *et al.* [19]. The testing dataset of KAIST contains 2,252 image pairs in which 797 pairs were captured during nighttime. The original annotations under the "reasonable" setting (pedestrians are larger 55 pixels and at least 50% visible) are used for performance evaluation [16].

**Implementation Details:** We apply the image-centric training scheme to generate mini-batches, which consist of 1 image and 120 randomly selected anchors. An anchor is considered as a positive sample if its Intersection-over-Union (IoU) ratio with one ground truth box is greater than 0.5, and otherwise negative. The first five convolutional layers in the each stream of TDNN (Conv1-V to Conv5-V in the visible stream and Conv1-T to Conv5-T in the thermal one) are initialized using the parameters of VGG-16 [33] deep convolutional neural networks pre-trained on the ImageNet dataset [32] in parallel. All the other convolutional layers and fully connected ones are initialized with a zero-mean Gaussian distribution with standard deviation (0.01). Deep neural networks are trained in the Caffe [17] framework with Stochastic Gradient Descent (SGD) [42] with a momentum of 0.9 and a weight decay of 0.0005 [20]. To avoid learning failures caused by exploding gradients [30], a threshold of 10 is used to clip the gradients.

**Evaluation Metrics:** We utilize the log-average miss rate (MR) [7] to evaluate the performance of multispectral pedestrian detection algorithms. A detected bounding box result is considered as a true positive if it can be successfully matched to a ground truth one (IoU exceed 50% [16]). Unmatched detected bounding boxes and unmatched ground truth ones are considered as false positives and false negatives, respectively. According to the method presented by Dollar *et al.* [7], detected bounding boxes matched to ignore ground truth ones do not be counted as true positives, as well unmatched ignore ground truth labels are not considered as false negatives. The MR is computed by averaging miss rate (false negative rate) at nine false positives per image (FPPI) rates evenly spaced in log-space from the range $10^{-2}$ to $10^0$ [16, 18, 19].

### 4.2. Evaluation on IFCNN

The illumination weighting mechanism provide an essential functionality in our proposed illumination-aware deep neural networks. We firstly evaluate whether IAFCNN can accurately calculate the illumination weights which provide critical information to balance outputs of illumination-aware sub-networks. We utilize the KAIST testing dataset, which contains multispectral images taken during daytime (1455 frames) and nighttime (797 frames), to evaluate the performance of IAFCNN. Given a pair of aligned visible and thermal images, IAFCNN will output a day illumination weight $\omega_d$. The illumination condition is correctly predicted if $\omega_d > 0.5$ for a daytime scene or $\omega_d < 0.5$ for a nighttime one. Moreover, we evaluate the performance of illumination prediction using feature maps extracted using visible channel (IFCNN-V) or thermal channel (IFCNN-T) individually, to investigate which channel provides the most reliable information to determine illumination condition of a scene. The architectures of IFCNN-V, IFCNN-T, and IFCNN are shown in Fig. 6 and their prediction accuracy are compared in Tab. 1.

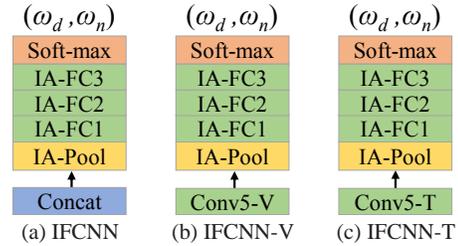

Figure 6: The architecture of IFCNN, IFCNN-V and IFCNN-T. Note that green boxes represent convolutional and fully connected layers, yellow boxes represent pooling layers, blue boxes represent fusion layers, and orange boxes represent soft-max layers. Best viewed in color.

Table 1: Accuracy of illumination prediction using IFCNN-V, IFCNN-T, and IFCNN.

|         | Daytime | Nighttime |
|---------|---------|-----------|
| IFCNN-V | 97.94%  | 97.11%    |
| IFCNN-T | 93.13%  | 94.48%    |
| IFCNN   | **98.35%** | **99.75%** |

It is observed that information from the visible channel can be used generate reliable illumination prediction for both daytime and nighttime scenes (daytime - 97.94% and nighttime - 97.11%). This result is reasonable as a human can easily determine it is a daytime scene or a nighttime one based on visual observation. Although thermal channel cannot be individually used for illumination prediction, it provides supplementary information to the visible channel to enhance the performance of illumination prediction. Through fusion of complementary information of visible and thermal channels, IFCNN compute more accurate illumination weights compared with IFCNN-V (using only visible images) or IFCNN-T (using only thermal images). The experimental results demonstrate that illumination condition of a scene can be robustly determined based on our proposed IFCNN by considering multispectral semantic features.

### 4.3. Evaluation of IATDNN

We further evaluate whether illuminate information can be utilized to boost the performance of multispectral pedestrian detector. Specifically, we compare the performances of TDNN and IATDNN . For fair comparison, information of semantic segmentation is not considered in both TDNN and IATDNN architectures. We combine the illumination loss term defined in Eq. 1 and detection loss term defined in Eq. 2 to jointly train IAFCNN and IATDNN, and use the detection loss term to train TDNN. TDNN model provides an effective framework to utilize two-stream feature maps (TSFM) for robust pedestrian detection [19]. However, it didn't differentiate human instances under day and night illumination conditions and use a common Con-Prov layer to generate detection results. In comparison, IATDNN apply the illumination weighting mechanism to adaptively combine outputs from multiple illumination-aware sub-networks (D-Cls, N-Cls, D-Reg, N-Reg) to generate the final detection results.

Table 2: MR of TDNN and IATDNN.

|  | All-day | Daytime | Nighttime |
| --- | --- | --- | --- |
| TDNN | 32.60% | 33.80% | 30.53% |
| IATDNN | **29.62%** | **30.30%** | **26.88%** |

Log-average miss rate (MR) is utlized as the evaluation metrics and the detection accuracies of IATDNN and TDNN are shown in Tab. 2. By considering the illumination information of a scene, IATDNN can significantly improve detection accuracy for both daytime and nighttime scenes. It also worth mentioning that such performance gain (TDNN 32.60% MR v.s. IATDNN 29.62% MR) is achieved at a cost of small computational overhead. Based on a single Titan X GPU, TDNN model takes 0.22s to process a paired of visible and thermal images (640×512 pixels) in KAIST dataset while IATDNN model needs 0.24s. More comparative results of computational efficiency are provided in Sec. 4.5. The experimental results demonstrate that illumination information can be robustly estimated based on multispectral data and further infused into multiple illumination-aware sub-networks for better learning of human-related feature maps to boost the performance of pedestrian detector.

### 4.4. Evaluation of IAMSS

We evaluate the performance gain by incorporating the semantic segmentation scheme with IATDNN. Here we compare the pedestrian detection using four different multispectral semantic segmentation models including MSS-F (feature-stage MSS), MSS (decision-stage MSS), IAMSS-F (illumination-aware feature-stage MSS) and IAMSS (illumination-aware decision-stage MSS). Architectures of these four models are shown in Figure 5. MSS models outputs a number of box-based segmentation masks, and such weakly annotated boxes provide additional information to enable the training of more distinctive features in IATDNN. The detection performance of IATDNN, IATDNN+MSS-F, IATDNN+MSS and IATDNN+IAMSS-F and IATDNN+IAMSS are compared in Tab. 3.

Table 3: Comparing the MR of TDNN+SS, IATDNN+SS, and IATDNN+IASS.

|  | All-day | Daytime | Nighttime |
| --- | --- | --- | --- |
| IATDNN | 29.62% | 30.30% | 26.88% |
| IATDNN+MSS-F | 29.17% | 29.92% | 26.96% |
| IATDNN+MSS | 27.21% | 27.56% | 25.57% |
| IATDNN+IASS-F | 28.51% | 28.98% | 27.52% |
| IATDNN+IAMSS | **26.37%** | **27.29%** | **24.41%** |

It is noticed that performance gains can generally be achieved through the joint training of pedestrian detection and semantic segmentation using all four different multispectral semantic segmentation models (except using IATDNN+MSS-F for nighttime scenes). The underlying principle is that semantic segmentation masks will provide additional supervision to facilitate the training of more sophisticated features for more robust pedestrian detection [2]. Another observation is that the choice of fusion scheme (feature-stage or decision-stage) will significantly affect the detection performance. Based on our evaluation, decision-stage multispectral semantic segmentation models (MSS and IA-MSS) performs much better the feature-stage models (MSS-F and IA-MSS-F). One possible explanation of this phenomenon is that late stage fusion strategy (e.g. decision-stage fusion) is more suitable to combine high-level segmentation results. Finding the optimal segmentation fusion strategy to process multispectral data will be our future research. Last but not least, performance of semantic segmentation can be boosted by considering illumination condition of the scene. Output of sub-networks are adaptively fused through the illumination weighting mechanism to generate more accurate segmentation results under various illumination conditions. Figure 7 shows comparative semantic segmentation results using four different MSS models. It is observed that semantic segmentation generated by IATDNN+IASS (using illumination) can more accurately cover small targets and suppress the background noise. More accurate segmentation can provide better supervision to train most distinctive human-related feature maps.

In Figure 8 we visualize the feature map of TDNN, IATDNN, and IATDNN+IAMSS to understand improvements gains achieved by different illumination-aware modules. We find that IATDNN generate more distinctive pedestrian features than TDNN by incorporating illumination information into multiple illumination-aware sub-networks for better learning of human-related feature maps. IATDNN+IASS can achieve further improvements through the segmenation infusion scheme in which illumination-aware visible and thermal semantic segmentation masks are used to supervise the training of feature maps.

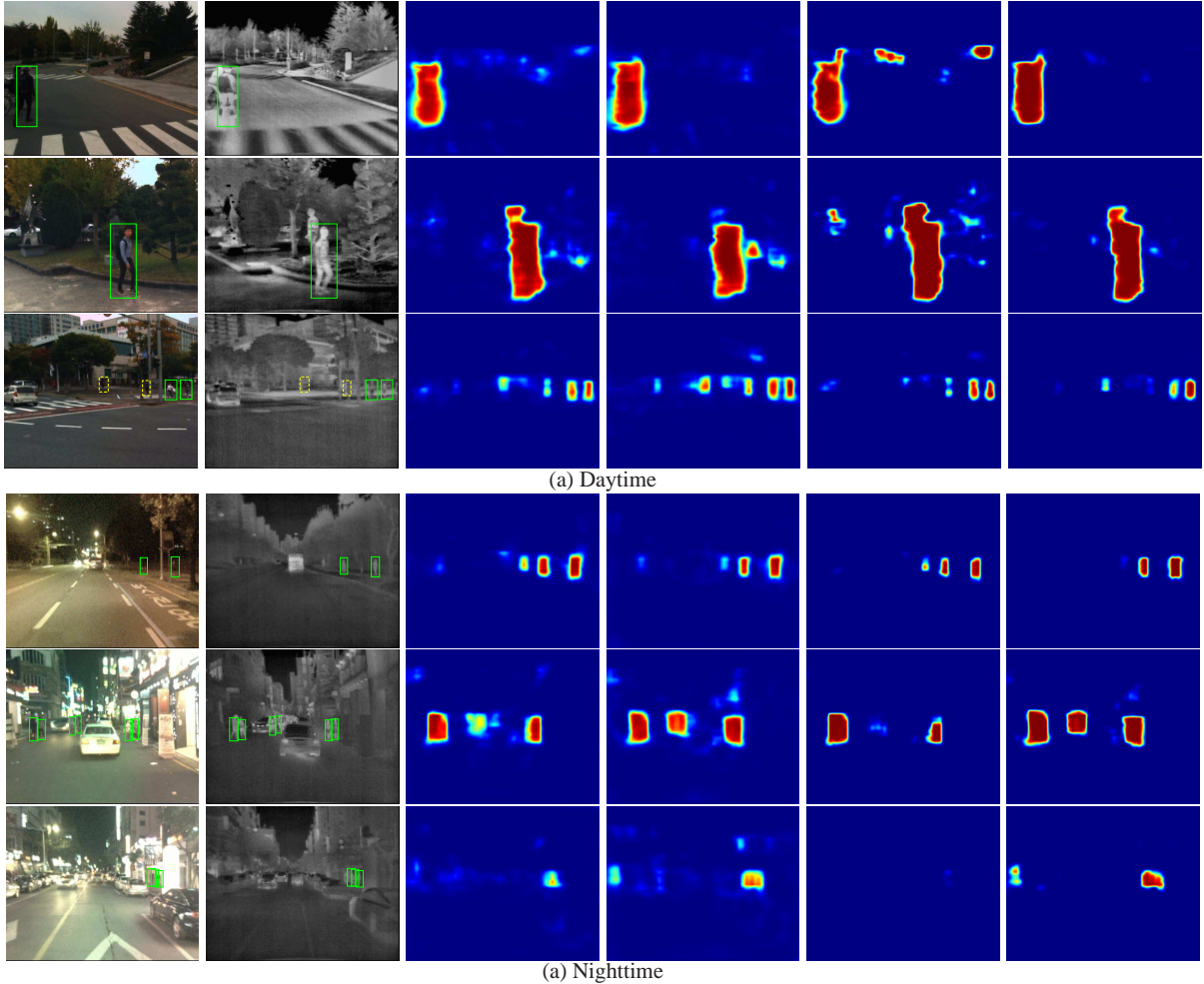

Figure 7: Examples of multispectral pedestrian semantic segmentation results generated using four different multispectral sematic segmentation models. The first two columns in (a) and (b) show the pictures of visible and thermal pedestrian instances respectively. The third to the sixth columns in (a) and (b) show the semantic segmentation generated from MSS-F, MSS, IAMSS-F and IAMSS respectively. Note that green bounding boxes (BBs) in solid line show positive labels, yellow BBs in dashed line show ignore ones. Best viewed in color.

### 4.5. Comparison with State-of-the-art Multispectral Pedestrian Detection Methods

Our proposed IATDNN and IATDNN+IASS are comparing with three other multispectral pedestrian detectors: ACF+T+THOG [16], Halfway Fusion [18] and Fusion RPN + BDT [19]. To compare detectors, we plot MR against FPPI (using log-log plots) by varying the threshold on detection confidence, as shown in Figure 9.

Our proposed IATDNN+IASS achieves an impressive 26.37% MR in all-day scenes. The performance gain is a relative improvement rate of 11% compared to the current state-of-the-art multispectral pedestrian detection method Fusion RPN + BDT (29.68%). Meanwhile, the performance of proposed detector surpass the state-of-the-art method in both daytime (27.29% vs. 30.51%) and nighttime (24.41% vs. 27.62%).

Furthermore, our proposed IATDNN, without using the semantic segmentation architecture, can achieve performance comparable to the state-of-art method (daytime: IATDNN (30.30%) vs. Fusion RPN + BDT (30.51%) and nighttime: IATDNN (26.88%) vs. Fusion RPN + BDT (27.62%)).

We visualize some detection results of the Fusion RPN + BDT and our proposed IATDNN and IATDNN+IASS in Figure 10. Comparing with the Fusion RPN + BDT, our proposed IATDNN and IATDNN+IASS is able to successfully detect most of the pedestrian instances under varying conditions of illumination. Meanwhile combining with illumination-aware semantic segmentation, the IATDNN+IASS reduces the false positives caused by double detections.

Furthermore, we compare the computing efficiency of IATDNN+IASS, IATDNN and TDNN with state-of-the-art methods as shown in Table 4. The efficiency of IATDNN+IASS surpasses the current state-of-the-art deep learning approaches for multispectral pedestrian detection by a large margin, with 0.25s/image vs. 0.40s/image on runtime. The architecture of Halfway Fusion is a combination of TDNN and Fast R-CNN [12]. It can be noticed that the Fast R-CNN model reduces the computing efficiency nearly by half. Meanwhile, the architecture of Fusion RPN + BDT is an ensemble of TDNN and boosted forest. We can observe that the boosting module is time consuming and increases the runtime by a factor of 3x. It is remarkable that our proposed illumination-aware weighting networks only have a little impact on network efficiency, with 0.25 vs. 0.22.

Table 4: Comparing the MR (all-day) and runtime performance of IATDNN+IASS with state-of-the-art methods. A single Titan X GPU is utilized to evaluate the computation efficiency. Note that DL represents deep learning and BF represents boosted forest [9] .

|  | MR(%) | Runtime (s) | Method |
| --- | --- | --- | --- |
| Halfway Fusion | 37.19 | 0.40 | DL |
| Fusion RPN+BDT | 29.68 | 0.80 | DL+BF |
| TDNN | 32.60 | 0.22 | DL |
| IATDNN | 29.62 | 0.24 | DL |
| IATDNN+IASS | 26.37 | 0.25 | DL |

## 5. Conclusion

In this paper, we propose a powerful multispectral pedestrian detector, which is based on multi-task learning of illumination-aware pedestrian detection and semantic segmentation. The illumination information encoded in multispectral images are utilized to compute the illumination-aware weights. We demonstrate that the weights can be accurately predicted by our designed illumination fully connected neural network (IFCNN). A novel illumination-aware weighting mechanism is developed to combine the day and night illumination sub-networks (pedestrian detection and semantic segmentation) together. Experiemntal results show that illumination-aware weighting mechanism provides an effective strategy to promote multispectral pedestrian detector. Moreover, we explore four different architectures for multispectral semantic segmentation and find illumination-aware decision-stage multispectral semantic segmentation generates the most reliable output. Experimental results on KAIST benchmark show that our proposed method outperforms state-of-the-art approaches and achieve more accurate pedestrian detection results using less runtime.

## 6. Acknowledgment

This research was supported by the National Natural Science Foundation of China (No.51575486, No. 51605428 and U1664264).

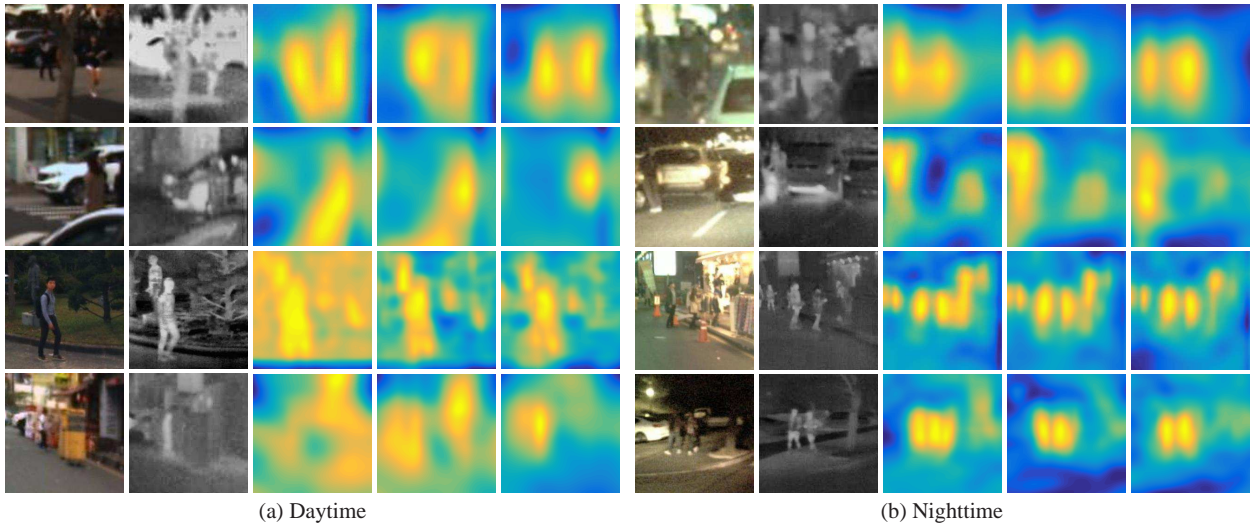

Figure 8: Examples of multispectral pedestrian feature maps which are promoted by illumination-aware mechanism captured in (a) daytime and (b) nighttime scenes. The first two columns in (a) and (b) show the pictures of visible and thermal pedestrian instances respectively. The third to the fifth columns in (a) and (b) show the feature map visualizations generated from TDNN, IATDNN and IATDNN+IASS respectively. Notice that the feature maps of multispectral pedestrian are improved by inserting our proposed two illumination-aware module IA (for classification and bounding box regression) and IASS (for generate multispectral semantic segmentation) progressively.

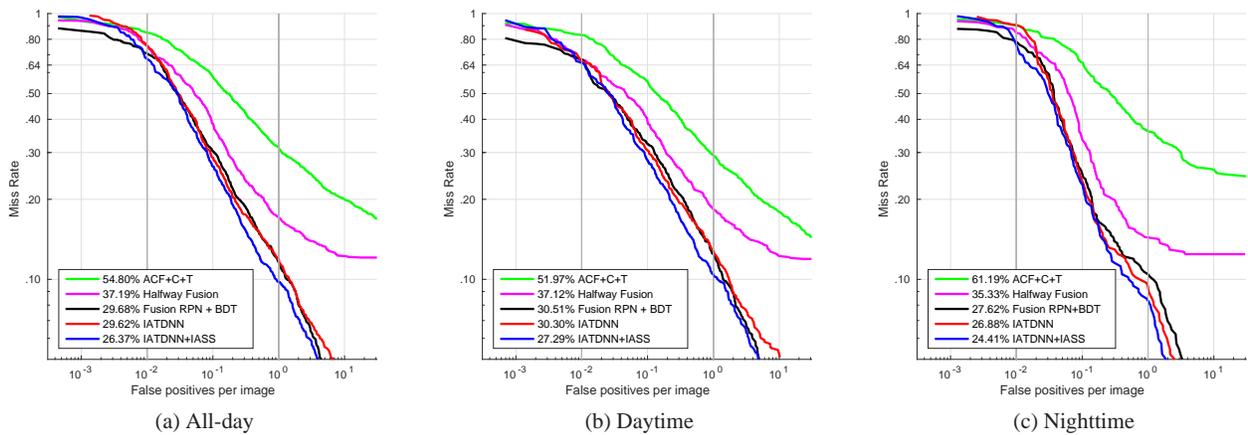

Figure 9: Comparisons on the KAIST test dataset under the "reasonable" setting during all-day (a), daytime (b), and nighttime (c) (legends indicate MR).

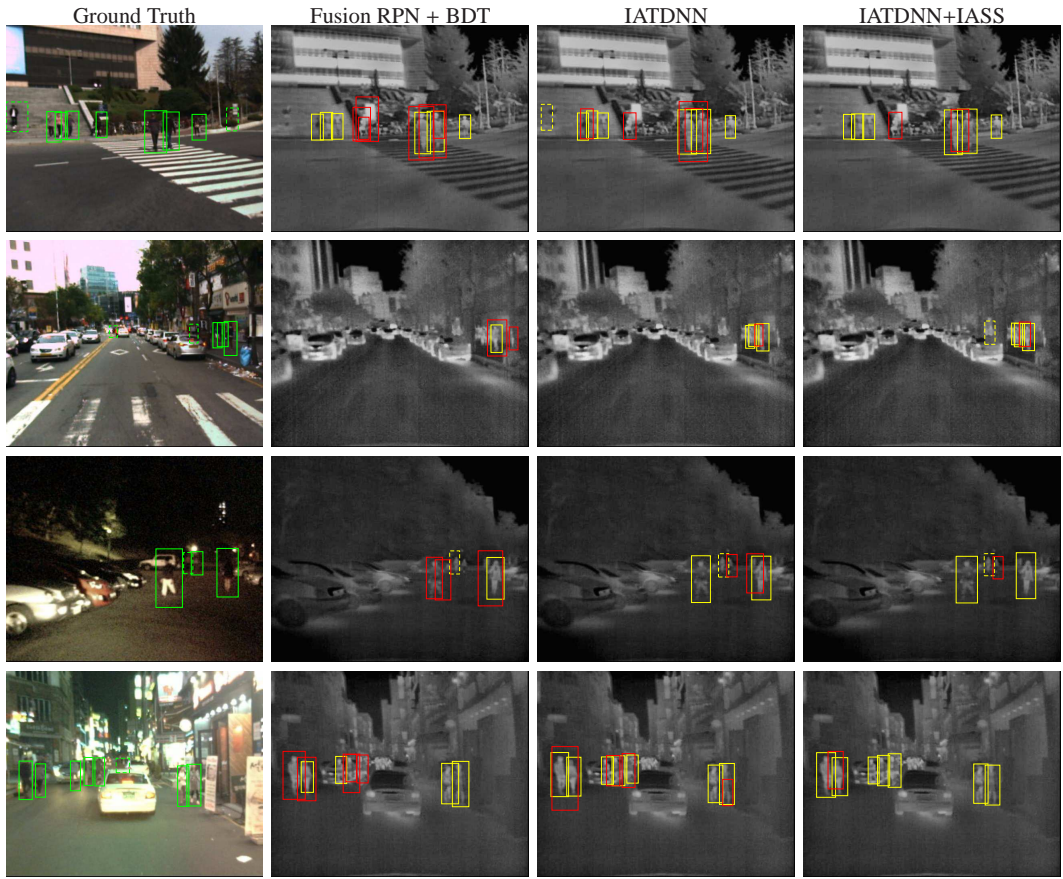

Figure 10: Comparison of pedestrian detection results with the current state-of-the-art approach (Fusion RPN + BDT). First column shows the input multispectral images with ground truth (displaying with visible channel) and the others show the detection results of Fusion RPN + BDT, IATDNN, and IATDNN+IASS (displaying with thermal channel). Note that green bounding boxes (BBs) in solid line show positive labels, green BBs in dashed line show ignore ones, yellow BBs in solid line show true positives, yellow BBs in dashed line show ignore detections, and red BBs show false positives. Best viewed in color.